\documentclass{article}
\usepackage{amsmath,amssymb}
\usepackage{multirow}
\usepackage{graphicx}
\usepackage{algorithm}
\usepackage{algorithmic}
\usepackage{color,soul}
\usepackage{booktabs} 
\usepackage{tabularx}
\usepackage{subcaption}
\usepackage{mwe}
\usepackage{commath}
\usepackage{verbatim}
\usepackage{mathtools}
\usepackage{cuted}
\usepackage{flushend}
\usepackage{arydshln}
\usepackage{multirow}

\usepackage{cite}
\usepackage{amsmath,amssymb,amsfonts}
\usepackage{algorithmic}

\usepackage{textcomp}
\usepackage{xcolor}
\usepackage{authblk}
\usepackage{times}  
\usepackage{helvet}  
\usepackage{courier}  
\usepackage{url}  
\usepackage{graphicx}  

\usepackage{mathtools}
\usepackage{amsmath,amssymb,amsfonts}

\usepackage{caption}
\usepackage{amsthm,dsfont}
\newcommand{\ignore}[1]{}
\usepackage{color}
\usepackage{authblk}
\usepackage{graphicx}
\usepackage{float} 
\usepackage{multirow}
\usepackage{amsfonts}
\usepackage{comment}
\usepackage{fancyhdr}
\usepackage{rotating}

\def\BibTeX{{\rm B\kern-.05em{\sc i\kern-.025em b}\kern-.08em
    T\kern-.1667em\lower.7ex\hbox{E}\kern-.125emX}}
\begin{document}

\title{On the Use of Diversity Mechanisms in Dynamic Constrained Continuous Optimization}

\author{Maryam Hasani-Shoreh}
\author{Frank Neumann}
\affil{Optimisation and Logistics, School of Computer Science,\\ The University of Adelaide, Adelaide, Australia}

\renewcommand\Authands{ and }
\maketitle

\begin{abstract}
Population diversity plays a key role in evolutionary algorithms that enables global exploration and avoids premature convergence. This is especially more crucial in dynamic optimization in which diversity can ensure that the population keeps track of the global optimum by adapting to the changing environment. Dynamic constrained optimization problems (DCOPs) have been the target for many researchers in recent years as they comprehend many of the current real-world problems. Regardless of the importance of diversity in dynamic optimization, there is not an extensive study investigating the effects of diversity promotion techniques in DCOPs so far.
To address this gap, this paper aims to investigate how the use of different diversity mechanisms may influence the behavior of algorithms in DCOPs. To achieve this goal, we apply and adapt the most common diversity promotion mechanisms for dynamic environments using differential evolution (DE) as our base algorithm. The results show that applying diversity techniques to solve DCOPs in most test cases lead to significant enhancement in the baseline algorithm in terms of modified offline error values.
 
\end{abstract}


\section{Introduction}
\vspace{-2mm}
\label{sec:intro}
Dynamic constrained optimization problems (DCOPs), in which the objective function or/and the constraints change over time, comprehend a variety of real world problems. 
An example of such problems is hydro-thermal scheduling~\cite{de2008plant}, in which the dynamism arises as the available resources or demand vary over time, or source identification problem~\cite{liu2008adaptive}, in which the information about the problem reveals gradually.
In these (so-called time-dependant or dynamic) optimization problems, the goal is to find and track the optimal solution of each instance of the dynamic problem given a limited computational budget. Another approach is to apply an independent optimization method to each problem instance, however, a more efficient approach tackles them in a dynamic manner, in which the algorithm detects and responds to the changes on-the-fly~\cite{Nguyen20121}. 
Mathematically, the objective is to find a solution vector ($\vec{x} \in \mathbb{R}^D $) at each time period $t$ such that: $\min_{\vec{x}\in F_t} f(\vec{x}, t)$
, where $f:S \rightarrow \mathbb{R}$ is a single objective function, and $t \in N^+$ is the current time period.  
$F_{t}=\{ \vec{x} \mid \vec{x} \in [L,U], g_i (\vec{x},t) \le 0,
h_j (\vec{x},t) = 0\}$,  is the feasible region at time $t$, where $L$ and $U$ are the boundaries of the search space; $g_i(x, t)$ and $h_j(x,t)$ are the linear $i$th inequality and $j$th equality constraints at time $t$, respectively.
To solve these problems, evolutionary algorithms (EAs) are commonly used as they have the ability to easily adapt to changing environments~\cite{Nguyen20121,Hasani-ShorehAB19}. 
In addition, they often provide good solutions to complex problems without a large design effort. 

For avoiding premature convergence that is a common problem in EAs a diverse population is needed. Otherwise there is no benefit of having a population; lack of diversity in population in the worst case, may lead the EA to behave like a local search algorithm, but with an additional overhead from maintaining many similar solutions~\cite{DirkDiversity}.
Premature convergence in dynamic environments pose more serious challenges to EAs as when they are converged, they cannot adapt well to the changes.
Indeed, having a diverse set of solutions in population helps to ensure the algorithm caters for changes in a dynamic environment.
In the literature of EAs, diversity has been found to have manifold positive effects. To name a few, it is highly beneficial for enhancing the global exploration capabilities of EAs. It enables crossover to work effectively, improves performance and robustness in dynamic optimization, and helps to search for the whole Pareto front for evolutionary multiobjective optimization~\cite{ishibuchi2010diversity}. In the related studies of runtime analysis~\cite{DirkDiversity}, diversity mechanisms proved to be highly effective for the considered problems, speeding up the optimization time by constant factors, polynomial factors, or even exponential factors. 

Diversity in EAs has been promoted through different approaches. A comprehensive classification is given in~\cite{vcrepinvsek2013exploration} that overally divides them into niching and non-niching approaches. Moreover, another classification is given based on the affected section of the algorithm: population-based, selection-based, crossover/mutation-based, fitness-based, and replacement-based. These approaches have been applied in many different classes of optimization problem so far including multi-objective~\cite{ishibuchi2010diversity}, multi-modal~\cite{yin1993fast}, constrained optimization~\cite{8477877}. To name a few of them in DCOPs based on the aforementioned classification include: mutation-based~\cite{cobb1990investigation}, replacement-based~\cite{grefenstette1992genetic} or population-based diversity mechanisms~\cite{Bu_2016, Goh_2009}. 

However, regardless of the manifold benefits of diversity in EAs and in particular in dynamic optimization, there is not an extensive study so far in DCOPs.
What makes study of diversity in this problem class important is that as diversity mechanisms spread the solutions over the search space, the constraint handling technique has tendency to guide the search toward feasible areas. The results of such study gives insight into the role of these opposing forces and their overall effect on algorithm's performance.
What we aim is to carry a survey study over commonly used explicit diversity promotion methods (we exclude implicit methods that are via parameter tuning or selection mechanisms) investigating their effects in DCOPs. Our comparison aims to reveal which diversity promotion technique work better in each specific problem characteristic and why.
Our investigations help to develop a better understanding of diversity role in DCOPs.

We choose differential evolution (DE) as our baseline algorithm which has showed competitive results in dynamic and constrained optimization problems~\cite{Ameca-AlducinHB18}. The presented results reveal applying the diversity promotion techniques enhanced algorithm performance significantly based on statistical test applied in modified offline error values.

The outline of the paper is as follows. Section~\ref{sec:intro}, gives an introduction and motivation to our work. Section~\ref{sec:Prim} introduces DE algorithm and diversity mechanisms. Experimental setup will be presented in Section~\ref{sec:ExperimentalSetup}. Results and discussion are reviewed in Section~\ref{sec:resultsAnalysis} and finally in Section~\ref{sec:conclusion} conclusions and future work are summarized.

\section{Preliminaries}
\label{sec:Prim}
\vspace{-2mm}
In this section, an overview of adapted differential evolution (DE) algorithm to solve DCOPs and diversity handling mechanisms are presented.

\subsection{DE algorithm for solving DCOPs}
\label{subsec:DE}
DE is a stochastic search algorithm that is simple, reliable and fast which showed competitive results in constrained and dynamic optimization~\cite{Ameca-AlducinHB18}. Each vector $\vec{x}_{i, G}$ in the current population (called at the moment of the reproduction as target vector) generates one trial vector $\vec{u}_{i, G}$ by using a mutant vector $\vec{v}_{i,G}$. The mutant vector is created applying $\vec{v}_{i,G}= \vec{x}_{r0,G} + F (\vec{x}_{r1,G} - \vec{x}_{r2,G})$,
where $\vec{x}_{r0,G}$, $\vec{x}_{r1,G}$, and $\vec{x}_{r2,G}$ are vectors chosen at random from the current population ($r0 \neq r1 \neq r2 \neq i$); $\vec{x}_{r0,G}$ is known as the base vector and $\vec{x}_{r1,G}$, and $\vec{x}_{r2,G}$ are the difference vectors and $F>0$ is a parameter called scale factor. The trial vector is created by the recombination of the target vector and mutant vector using a crossover probability $CR \in [0,1]$. 

For the constraint handling technique feasibility rules~\cite{deb2000efficient} is applied. 

In addition to constraint handling techniques, the algorithms in DCOPs need a mechanism to detect the environment changes.
In the related literature, re-evaluation of the solutions is the most common change-detection approach~\cite{Nguyen20121}. The algorithm regularly re-evaluates specific solutions (for us the first and the middle individual of the population) to detect changes in their function values or/and constraints. 
If a change is detected, then all the population is re-evaluated to avoid obsolete information.

\subsection{Diversity Maintenance Techniques}
\label{subsec:divmech}
In this section diversity handling mechanisms are reviewed. Among the many popular niching methods like fitness sharing, clearing and species-based, we used standard crowding in this work. The reason for excluding other niching methods is that they are originally designed and applied in multi-modal functions. An extensive separate study is needed to apply these methods with moving peak benchmark (that is designed for testing multi-modal optimization in DCOPs) to investigate the methods thoroughly. 

\textbf{Chaos local search:}
Chaos is a nature phenomenon characterized by randomness and sensibility to initial conditions. Due to those attributes, chaos has been implemented with success in local searches~\cite{jia2011effective}, which is the case for this mechanism for promoting diversity. In our case the chaos only affects the best solution at each iteration. We applied an adaptive dynamic search length that is triggered with change detection.

\textbf{Crowding:}
 Among the many niching methods in literature, we choose the standard crowding method\cite{sareni1998fitness}. In this method, similar individuals in the population are avoided, creating genotypic diversity. Instead of competition between the offspring and the parents, the offspring competes with the individual with lowest Euclidean distance.

\textbf{Fitness diversity:}
While the other methods focus is on creating genotypic diversity, this method creates phenotypic diversity by avoiding individuals with close fitness values. The offspring in this method competes with the individual with closest fitness value~\cite{hutter2006fitness}.

\textbf{No diversity mechanism:}
This method is a base DE algorithm with feasibility rules~\cite{deb2000efficient} used as its constraint handling technique and no explicit method is used as its diversity promotion technique. Note that all the other methods use feasibility rules as their constraint handling mechanism.

\textbf{Opposition:}
This mechanism is based in the estimation of the symmetric opposites of individuals in the population, since it leads to find new positions closer to the problem optimum~\cite{rahnamayan2008opposition}. The authors claim that evaluating opposites, when solving a problem without a priori knowledge with several dimensions, helps in finding fitter individuals.
Purely random re-sampling or selection of solutions from a given population has the chance of visiting or even revisiting unproductive regions of the search space. 

\textbf{Random immigrants:}
This infusion technique replaces certain number of individuals (defined by a parameter called replacement rate) with random solutions in the population, so as to assure continuous exploration~\cite{grefenstette1992genetic}. 

\section{Experimental setup}
\vspace{-2mm}
\label{sec:ExperimentalSetup}
In this section, the applied performance measures and the test problems are reviewed.
\vspace{-2mm}
\subsection{Performance measures}
\label{subsec:measures}
The first two performance measures calculate the precision of solutions compared to optimal values. The third and fourth measures represent speed of convergence and the success rate for each algorithm to reach to a precision from the optimum. Finally, the last measure calculates diversity of solutions in population.

\textbf{Modified offline error ($MOF$)}
is defined as: $\frac{1}{G_{max}} \sum_{G = 1}^{G_{max}} (|f(\vec{x}^*,t) - f(\vec{x}_{best,G},t)|)
\label{eq:offlineerror}$ that represents the average of the sum of errors in each generation divided by the total number of generations~\cite{nguyen2012continuous}.

Where $G_{max}$ is the maximum generation, $f(\vec{x}^*,t)$ is the global optimum at current time $t$, and $f(\vec{x}_{best,G},t)$ represents the best solution found so far at generation $G$ at current time $t$.

\textbf{Tracking error ($TE$)}
is defined as: $\frac{1}{t_{end}-t_0} \sum_{t = t_0}^{t_{end}} |f(\vec{x}^*,t) - f(\vec{x}_{best},t)|$ and reflects the overall error at the end of each change period.

\textbf{Number of fitness evaluations ($NFE$)}
needed at each time to reach to an $\epsilon$-precision from the global optimum are averaged over all the times for this measure. The termination criteria is to find a value smaller than the $\epsilon$-level from the global optimum (value to reach ($VTR$)) before reaching to the next change.

\begin{equation}
\begin{array}{l}
NFE= \frac{1}{t_{end}-t_0} \sum_{t = t_0}^{t_{end}} NFE_t\\\\

VTR=\frac{|f(\vec{x}^*,t) - f(\vec{x}_{best,G},t)|}{|f(\vec{x}^*,t)|}
\end{array}
\end{equation}

\textbf{Success rate ($SR$)} calculates the percentage of the number of times each algorithm is successful to reach to $\epsilon$-precision from the global optimum ($VTR$) over all time scale.

\textbf{Diversity:}
Diversity measures differences among individuals at distinct levels; genotypic: considers individuals position within the search space or  phenotypic: evaluate populations fitness distribution. 
We choose a genotypic measure as its more common in the literature. For this purpose we measure relative standard deviation of the population (known as coefficient of variation): $CV=\frac{\sigma}{\mu}$ at each generation, where $\sigma$ is the standard deviation and $\mu$ is the mean of the population.

\subsection{Test problems}
Our algorithms were tested on two benchmarks for DCOPs~\cite{Benchmark,Bu_2016}, which contains 22 problems in total. The first 18 test cases are from~\cite{Benchmark} that captures different characteristics of DCOPs like multiple disconnected feasible regions, gradually moving feasible regions and global optimum switching between different feasible regions. The last 4 test cases are from~\cite{Bu_2016} in which Bu used a parameter in the original test cases in~\cite{Benchmark} that controls the size and the number of disconnected feasible regions.
In the experiments, medium severity is chosen for the objective function ($k=0.5$) and the constraints ($S=20$). The other parameters are: frequency of change ($f_c$)=1000, runs=30 and the number of considered times for dynamic perspective of the algorithm $5/k$ ($k=0.5$).
Parameters of DE are chosen as $n_p=20$, $CR=0.2$,  $F$ is a random number in $[0.2,0.8]$, and rand/1/bin is the chosen variant of DE~\cite{Ameca-AlducinHB18}. 

\section{Results and Discussion}
\vspace{-2mm}
\label{sec:resultsAnalysis}
In this section first the results for diversity measure is reviewed and then the results for $MOF$, $TE$, $SR$, and $NFE$ values will be discussed.
\subsection{Diversity results}
\begin{figure*}[t]
        \centering          \includegraphics[width=0.9\textwidth, height=5cm]{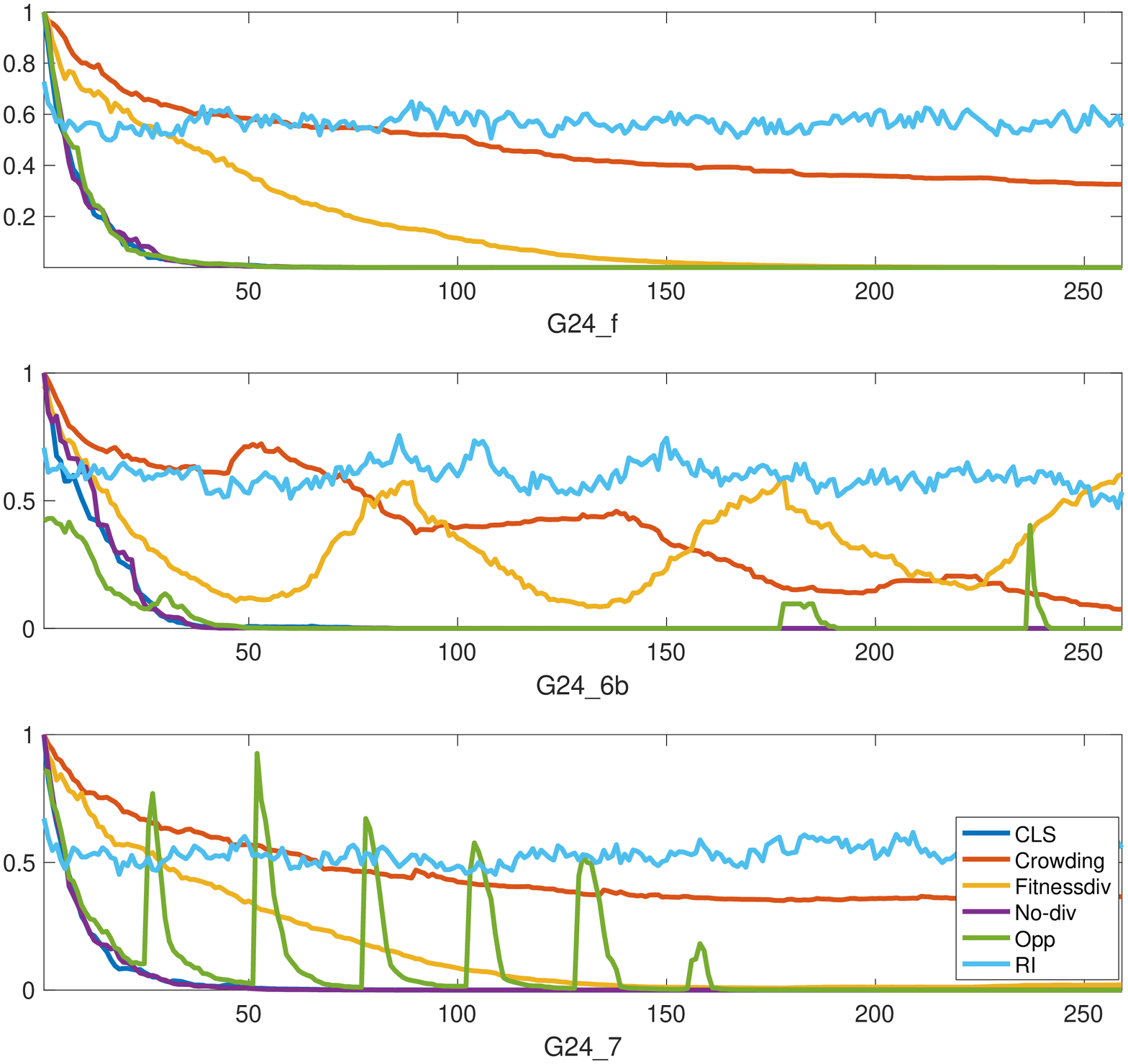} 
        \caption{\scriptsize  Y-axis: Diversity score (coefficient of variation of population), X-axis: Generations}
        \label{fig:diversity}
        \vspace{-4mm}
    \end{figure*}

Figure~\ref{fig:diversity} illustrates the results for coefficient of variation of population (a measure for considering diversity explained in Section~\ref{subsec:measures}) for different methods per generation. Three functions have been opted for plots considering a range of different characteristics. Notice that the generations are not equal for all methods as the frequency of changes is mapped with the number of fitness evaluations, and some methods like $Opp$ and $RI$ use different number of fitness evaluations per generations compared to the other methods. $Opp$ has almost half the number of generations when a change happens and $RI$ is different from the other methods within a range (based on what the replacement rate is).

In general $RI$ shows almost the same trajectory regardless of the test case. It starts with maximum diversity around 0.6 and remains with a minor drop through the last generations. The lack of convergence in this method is due to random individuals inserted in the population at each iteration keeping diversity at a consistent level.
For $No-div$ and $CLS$ also an identical behaviour is observed regardless of the test case. They both start with a high value for diversity measure equal to 1 and converge to near zero after 45 generations which represents the number of generations in which the first change happens. Thus as diversity measure shows these two methods are not able to promote any diversity in population after they converge in the first change.
As explained in Section~\ref{subsec:DE}, the way the tested algorithms react to changes is through re-evaluation of the solutions. For the other methods due to the applied diversity promotion technique, they can diverge faster leading to higher $MOF$ values overall (is discussed in next section). But for these two methods finding new optimum after a change is very slow. This is because DE relies on differential vectors to maintain diversity, which are dependent on the population's diversity itself. So without increasing diversity extrinsically, diversity will remain low in them leading their inability to track new optimum.

For $Opp$ the diversity depends weather the opposite individual is accepted in the population or not. For test cases with smaller feasible regions, the diversity is low as the opposite individual is infeasible and hence it is rejected. As if it is accepted there is more diversity in the population otherwise this method behaves like $\text{No-div}$ technique.
For G24\_7, the feasible area shrinks from 44.61\% to 7.29\% over time. This explains the behavior of $Opp$ in which at each change, the diversity increases sharply and then reduces gradually until next change happens. This pattern is repeated until around generation 160. From this generation afterwards, it loses its diversity as all of the opposite individuals fail to be chosen in the selection process (due to small feasible region). 

$Crowding$ in all test cases is able to maintain high diversity over many generations, thus keeping its effectiveness in responding to dynamic environment.
This method shows higher diversity near to 0.9 at the first generations and linger around o.4 until termination for functions G24\_f and G24\_7. For function G24\_6b its behaviour is a bit different as this test case has a special characteristic. For this test case, the objective function changes over time causing the global optimum to switch between two corners of the search space at each change step. This characteristic in this function can attribute the oscillated behaviour of diversity in $Fitnessdiv$ method. By starting the first time before a change happen at generation 45, the diversity decreases. When the change happens as the new optima is in the other boundary of the search space, so the solutions must diverge again gradually to reach to the new optimum on the other corner of the search space until reaches to the next change in the environment and the same pattern repeats.
For the other two test cases, $Fitnessdiv$ showed a similar trend to $Crowding$ but with lower diversity. Another difference is after 150 generations it looses its diversity. However, $Crowding$ still keeps its diversity at around 0.4 until the end of generations.
In general, as our selected measure of diversity is based on 
genotypic level, $Crowding$ that is based of promoting diversity in genotypic level has got higher scores. 
    
\subsection{Statistical results}
The results obtained for the compared algorithms using $MOF$ values are summarized in Table ~\ref{tab:offline error}. Furthermore, for the statistical validation, the 95\%-confidence Kruskal-Wallis (KW) test and the Bonferroni post-hoc test, as suggested in \cite{Derrac20113} are presented (see Table \ref{tab:KWtest}). 

Based on Tables~\ref{tab:offline error} and~\ref{tab:KWtest}, $MOF$ shows significantly superior results for almost all of the methods compared to the base algorithm, $No-div$, for most of the test cases. Among them, $Crowding$ has the highest frequencies of wins over the base algorithm compared to other methods.
The difference of outperformance of $Crowding$ compared to other methods is more significant in test cases by dynamic objective functions. This shows the other methods decrease in their performance dealing with dynamic objective functions compared to static ones, leading to bigger difference in dynamic cases. 

On the contrary the method that is very similar to $No-div$ method is $CLS$. As diversity results showed, this method is not increasing diversity that much, as it is more like a local search. This explains its superior performance compared to other methods in fixed test cases (G24\_3f, G24\_f). But it can not promote diversity that much (as only best solution at each iteration changes) and its inability to react to changes explains its higher $MOF$ values.

\begin{figure*}[t]
    \centering          \includegraphics[width=1.05\textwidth, height=7.5cm]{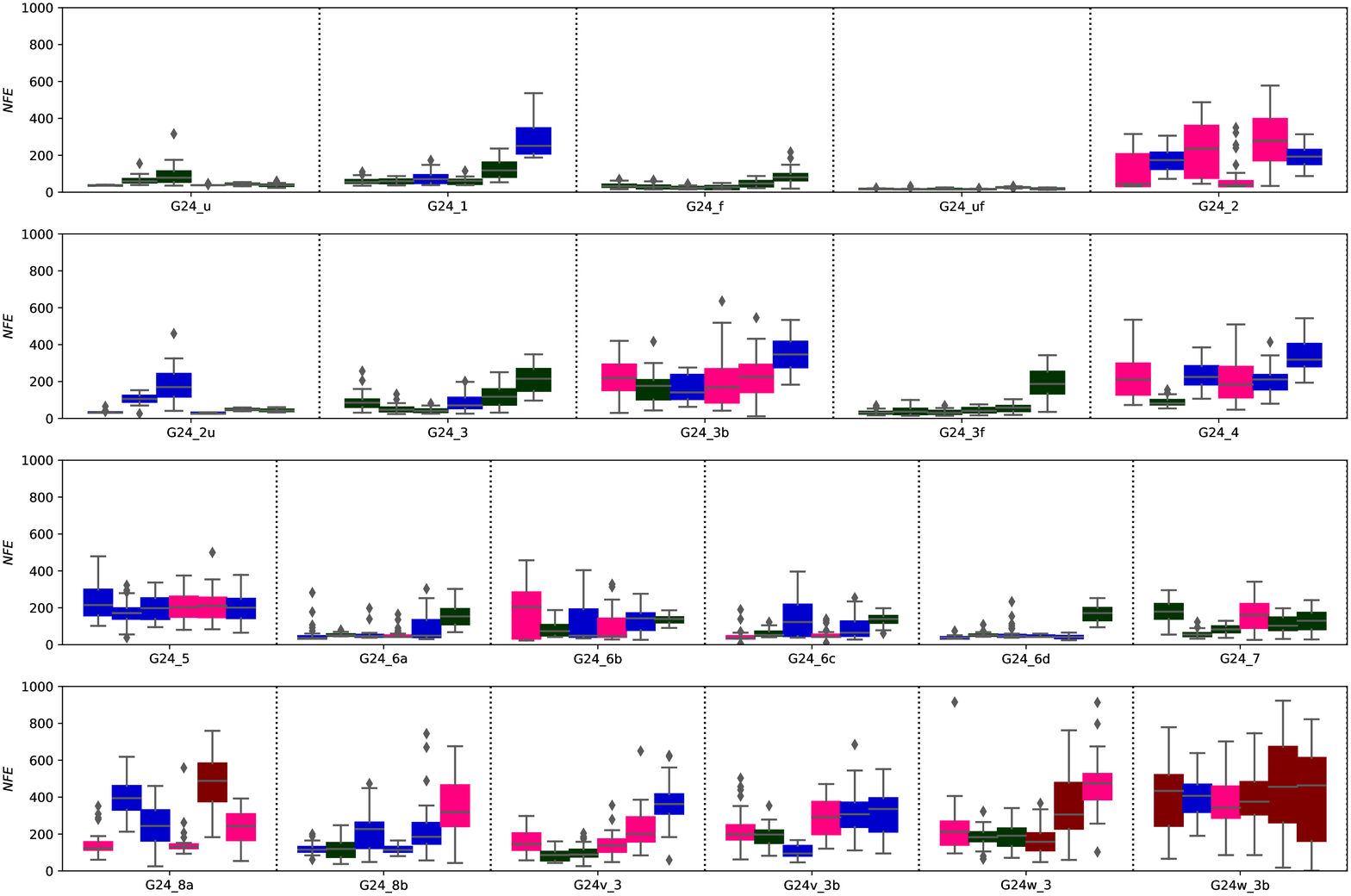}  
    \caption[]
    {\scriptsize  Boxplot of $NFE$ values for $\epsilon=10\% VTR$, color-coded with $SR$ values: dark-red: $SR<20\%$, purple: $SR \in [20\%-50\%]$, blue: $SR \in [50\%-80\%]$, dark-green: $SR>80\%$.\\
    From left to right $1=CLS$, $2=Crowding$, $3=Fitnessdiv$, $4=No-div$, $5=Opp$, and $6=RI$.} 
    \label{fig:boxplotfeval10percent}
    \vspace{-5mm}
\end{figure*}

$RI$ preformed worse than the base algorithm for the problems with small feasible area (like G24\_3f, G24\_f, G24v\_3, G24w\_3 ) as there is a high probability that the inserted solutions are infeasible and hence they can not compete with the current best solution based on the applied constraint handling technique. On the contrary, for unconstrained ones and the ones with large feasible areas showed the best results (like G24\_u, G24\_2, G24\_2u, G24\_6b, G24\_8a). So generally, although the diversity of population is increased by inserting new solutions to the population, but because they are not feasible solutions they are not that effective. In addition, in this method the algorithm spends some more fitness evaluations (to the extent of replacement rate) compared to the base algorithm at each generation. Thus the algorithm will have less computation budget for the evolution process itself as the changes happen after known number of fitness evaluations.
In this benchmark there are some pairs of test cases that are used to test one behaviour of the algorithms like their abilities to handle constraints in the boundaries or the cases with disconnected feasible regions versus non-disconnected feasible area. 
Comparing the pairs of test cases with optima in constraint or search boundary over optima not in constraint or search boundary, with $No-div$ the optima in constraint boundary has got better results (G24\_1, G24\_2), (G24\_4, G24\_5), (G24\_8b, G24\_8a). While for other methods the trend is similar, for $RI$ the trend is the other way, meaning with increasing diversity by inserting random solutions in $RI$, algorithm enhances its ability for finding optima which is not in constraint boundary.
The pairs of fixed constraints versus dynamic constraints (G24\_f, G24\_7), (G24\_3f, G24\_3) show significant decrease in $MOF$ values for dynamic constraints as it gets harder for the methods to deal with dynamism. The only exception is the behaviour of $RI$ method in which $MOF$ values are almost the same in these two cases and this is attributed with stochastic nature of this method.
The pairs of test cases for observing the effect of connected feasible region versus disconnected feasible region (G24\_6b, G24\_6a), (G24\_6b, G24\_6d) do not show any trend in the results. As this behaviour depends to the constraint handling technique of algorithms to a greater extent, that is similar in our case.
$Opp$ showed similar performance to the base algorithm in most of the cases with some exceptions. As the authors claim this method is to enhance convergence speed compared to base DE algorithm and is more effective in higher dimension problems. Our case is a two dimension, it is recommended to test this algorithm with higher dimension to see its effects in DCOPs.
Due to space limitation, we discarded to bring the results of $TE$ values. But the results of this measure also proof $Crowding$ has the best performance over the other methods in most of the test cases. In addition for two test cases (G24\_2, G24\_8a) that $RI$ showed better results in $MOF$ values, based on $TE$ values $Crowding$ has better performance meaning $Crowding$ end-up to closer values to optimum at the end of the change periods. For G24\_uf all the methods except $RI$ reach to optimum values as the $TE$ values are zero. Also for G24\_u, $Crowding$ and $RI$ reach to optimum at all times. 
Figure~\ref{fig:boxplotfeval10percent} shows the distribution of $NFE$ values in box-plots. Boxes represent the values obtained in the central 25\%-75\%, while the line inside the box shows the median values. Whiskers depicts highest and lowest values within interquartile range and dots show outliers.
The figure is color-coded based on $SR$ values. The dark-red color shows $SR$ values lower than 20\%, the purple color belongs to $SR \in [20\%-50\%]$, the blue color is for $SR \in [50\%-80\%]$, and finally the dark-green belongs to $SR$ values above 80\%.
In general for the test cases with fixed characteristics (G24\_f, G24\_uf, G24\_3f) or unconstrained cases (G24\_u, G24\_2u) in most cases the algorithms manage to reach to $VTR$ values in lower $NFE$ and higher $SR$ values.
For the rest of the test cases, there is this general observation that shows regardless of the test case, $Crowding$ and $Fitnessdiv$ are more successful based on $SR$ values that is easily observable based on stages that we defined and color-coded. Based on the colors, these two methods are usually one stage higher than the other methods in $SR$ values.
Based on the characteristics of some functions, the distribution of $NFE$ show quiet a high standard deviation in results.
These cases show less reliability in algorithms behaviour as in each run they achieve the $VTR$ in different $NFE$ values.
Those belong to test cases with either smaller feasible areas or the test cases with specific characteristics (G24\_2, G24\_5). In these two test cases in some change periods the landscape is either a plateau or contains infinite number of optimum.
In some cases the algorithms are almost unable to reach $VTR$ in the given number of evaluations. The cases are colored with dark-red based of $SR$ values. Figure~\ref{fig:boxplotfeval10percent} for $RI$ shows in unconstrained cases it reaches to $10\% VTR$ values with 100\% of $SR$ in almost of the cases but with high $NFE$ values. In addition, for the functions with small feasible areas the results show a high number of standard deviation. Indeed, this was expected based on the random nature of this method. So depending on the random solutions inserted, in some cases they can manage to track the optima and in the others they are unable to do so. In general $RI$ is the most different method that has different trend compared to other methods in almost all the functions.

\subsection{Discussions}
\vspace{-1mm}
Overall, results show $Crowding$ that has higher diversity in population could reach to better $MOF$ values. It also shows higher speed and frequency of reaching to $10\%VTR$ values due to the results of $SR$ and $NFE$ values. Statistical test shows (Table~\ref{tab:KWtest}) this method not only have significantly better results compared to base algorithm ($No-div$), but also compared to other methods in most of the test cases. While this method shows competitive results regardless of the test case, success of some methods highly depend on the tested problem.
In this method high diversity in population is created by avoiding genotypic similar individuals and without the need for extra fitness evaluations. While for some other methods like $RI$ high diversity in population is achieved with the need for extra evaluations and the new solutions are inserted randomly. While the random insertion of solutions is often successful in unconstrained problems, in constrained problems it is dependant to the feasibility status of the new solutions and the applied constraint handling method.
Indeed the constraint handling technique forces the solutions toward feasible areas avoiding infeasible areas while diversity mechanisms tend to have solutions spread over the search space. Of course the severity of the competition depends to both opposing forces: diversity promotion technique and constraint handling technique applied.
So depending to the methods applied, the created solution may not be accepted (rejected by constraint handling method) to be inserted to the population like in $Opp$, leading to low diversity in constrained problems; Or the solutions are inserted in the population increasing diversity (like in $RI$) but unable to compete with the best solution.
Thus one solution is applying adaptive constraint handling mechanisms. They can be designed in a way that at the first generations after a change, they increase their threshold (more relaxed with constraint violation) in order to allow diversity mechanisms explore the whole search space and then decrease to conduct the search toward feasible areas. The other suggestion is to use repair methods~\cite{ameca2018use} as constraint handling technique, in which if the created solution by diversity handling mechanism is infeasible but a good solution (in terms of fitness function), preserve it in the population by repairing it.
For more elaborate investigations on the roles of these two forces over each other and the overal algorithm performance, a measure that shows percentages of feasible and infeasible solutions that are selected for next generation can be helpful. As this measure can imply diversity handling mechanism ability to how it will balance exploiting feasible regions while exploring infeasible regions.

As the results of $Fitnessdiv$ are not as promising as $Crowding$ in the problem types in this benchmark, this can be concluded as phynotypic diversity has less effectiveness as gynotypic diversity.
For $CLS$, although an adaptive approach is used in such a way that at the first generations after a change, the local search length is large and gradually reduces; But as this local search is only applied to best solution at each generation, it is not enough to promote diversity as the results show. One solution is to do chaos local search randomly for some other individuals of the population besides to the best solution to increase diversity more. $RI$ showed less reliability in $NFE$ and $SR$ values (based on high standard deviation in results) and its worse performance for $MOF$ values for problems with small feasible area. Conversely, this method is highly suggested to be used for unconstrained problems as it showed best results for these cases.


\begin{table*}[t]
\renewcommand{\arraystretch}{1.0}
\centering
\caption{\scriptsize Average and standard deviation of $MOF$ values over 30 runs. Best results are remarked in boldface.}
\label{tab:offline error}
	\scalebox{0.65}{
\begin{tabular}{c|cccccc}
\hline\multirow{2}{*}{\textbf{Algorithms}}&\multicolumn{6}{c}{\textbf{Functions}}\\ &
     \textbf{G24\_u}&\textbf{G24\_1}&\textbf{G24\_f}&\textbf{G24\_uf}&\textbf{G24\_2}&\textbf{G24\_2u}\\\hline 
     CLS&0.4331($\pm$0.022)&0.5658($\pm$0.014)&\textbf{0.0294($\pm$0.012)}&0.0027($\pm$0.002)&0.7965($\pm$0.085)&0.3125($\pm$0.058)
     \\Crowding&0.0576($\pm$0.027)&\textbf{0.0823($\pm$0.025)}&0.0698($\pm$0.029)&0.0033($\pm$0.004)&0.2030($\pm$0.049)&0.2488($\pm$0.247)
     \\Fitnessdiv&0.2084($\pm$0.161)&0.5804($\pm$0.014)&0.0394($\pm$0.014)&0.0046($\pm$0.004)&0.9292($\pm$0.202)&0.8327($\pm$0.293)
     \\No-div&0.6215($\pm$0.002)&0.5737($\pm$0.014)&0.0332($\pm$0.014)&0.0026($\pm$0.002)&1.5727($\pm$0.091)&1.5208($\pm$0.002)
     \\Opp&0.6118($\pm$0.013)&0.5364($\pm$0.080)&0.0446($\pm$0.019)&\textbf{0.0020($\pm$0.002)}&1.3765($\pm$0.076)&0.0299($\pm$0.015)
     \\RI&\textbf{0.0344($\pm$0.017)}&0.5183($\pm$0.044)&0.4356($\pm$0.040)&0.0031($\pm$0.003)&\textbf{0.1935($\pm$0.029)}&\textbf{0.0292($\pm$0.011)}
     \\\hline &
     \textbf{G24\_3}&\textbf{G24\_3b}&\textbf{G24\_3f}&\textbf{G24\_4}&\textbf{G24\_5}&\textbf{G24\_6a}\\\hline 
     CLS&0.2433($\pm$0.058)&0.7861($\pm$0.108)&\textbf{0.0263($\pm$0.014)}&0.8449($\pm$0.109)&0.7077($\pm$0.099)&1.9470($\pm$0.146)
     \\Crowding&0.1618($\pm$0.039)&\textbf{0.1556($\pm$0.023)}&0.0471($\pm$0.017)&\textbf{0.1204($\pm$0.017)}&\textbf{0.1628($\pm$0.033)}&\textbf{0.0734($\pm$0.022)}
     \\Fitnessdiv&\textbf{0.0745($\pm$0.013)}&0.5126($\pm$0.112)&0.0314($\pm$0.009)&0.6216($\pm$0.052)&0.8183($\pm$0.134)&1.8504($\pm$0.131)
     \\No-div&0.3619($\pm$0.152)&0.9496($\pm$0.192)&0.0269($\pm$0.011)&0.6756($\pm$0.073)&1.2418($\pm$0.094)&2.1163($\pm$0.434)
     \\Opp&0.3036($\pm$0.117)&0.6724($\pm$0.180)&0.0460($\pm$0.020)&0.3271($\pm$0.070)&0.9553($\pm$0.041)&1.5525($\pm$0.546)
     \\RI&0.4339($\pm$0.038)&0.5102($\pm$0.037)&0.4079($\pm$0.040)&0.5129($\pm$0.043)&0.2170($\pm$0.034)&0.2753($\pm$0.075)
     \\\hline &
     \textbf{G24\_6b}&\textbf{G24\_6c}&\textbf{G24\_6d}&\textbf{G24\_7}&\textbf{G24\_8a}&\textbf{G24\_8b}\\\hline 
     CLS&0.9702($\pm$0.397)&1.7536($\pm$0.458)&0.6860($\pm$0.013)&0.3440($\pm$0.094)&0.7243($\pm$0.031)&0.6747($\pm$0.028)
     \\Crowding&0.2554($\pm$0.064)&\textbf{0.0993($\pm$0.050)}&\textbf{0.0775($\pm$0.015)}&\textbf{0.1475($\pm$0.026)}&0.5850($\pm$0.030)&\textbf{0.1864($\pm$0.039)}
     \\Fitnessdiv&1.1505($\pm$0.377)&0.9252($\pm$0.523)&0.5438($\pm$0.061)&0.1614($\pm$0.032)&0.5200($\pm$0.056)&0.7040($\pm$0.069)
     \\No-div&1.7479($\pm$0.789)&1.9300($\pm$0.646)&0.7887($\pm$0.010)&0.3442($\pm$0.162)&1.1064($\pm$0.013)&0.7265($\pm$0.007)
     \\Opp&0.8291($\pm$0.624)&0.8088($\pm$0.696)&0.7756($\pm$0.012)&0.2025($\pm$0.148)&1.1914($\pm$0.050)&0.7510($\pm$0.032)
     \\RI&\textbf{0.1876($\pm$0.023)}&0.1882($\pm$0.025)&0.2320($\pm$0.039)&0.3904($\pm$0.044)&\textbf{0.4755($\pm$0.031)}&0.5557($\pm$0.041)
     \\\hline &
     \textbf{G24v\_3}&\textbf{G24v\_3b}&\textbf{G24w\_3}&\textbf{G24w\_3b}\\\hline 
     CLS&0.5923($\pm$0.507)&0.6941($\pm$0.205)&1.0406($\pm$0.503)&1.2456($\pm$0.161)
     \\Crowding&\textbf{0.2529($\pm$0.081)}&\textbf{0.2214($\pm$0.057)}&0.5503($\pm$0.198)&\textbf{0.6458($\pm$0.141)}
     \\Fitnessdiv&0.2943($\pm$0.147)&0.4219($\pm$0.051)&\textbf{0.4784($\pm$0.161)}&0.6905($\pm$0.186)
     \\No-div&0.8776($\pm$0.701)&0.9734($\pm$0.222)&1.1860($\pm$0.467)&1.2710($\pm$0.215)
     \\Opp&0.7725($\pm$0.719)&0.6778($\pm$0.132)&1.1778($\pm$0.371)&1.1999($\pm$0.213)
     \\RI&1.4512($\pm$0.101)&0.8169($\pm$0.097)&1.2991($\pm$0.091)&1.2263($\pm$0.085)
     \\\hline
     \end{tabular}
}
\vspace{-2mm}
\end{table*}


\begin{table*}[t]

\renewcommand{\arraystretch}{1.0}
\centering

\caption{\scriptsize The 95\%-confidence Kruskal-Wallis (KW) test and the Bonferroni post-hoc test on the $MOF$ values in Table \ref{tab:offline error}. The compared variants are denoted as:  $1=CLS$, $2=Crowding$, $3=Fitnessdiv$, $4=No-div$, $5=Opp$, and $6=RI$. }
\label{tab:KWtest}
	\scalebox{0.6}{
\begin{tabular}{l|l}
\hline\textbf{Functions}& \textbf{Statistical Test}\\\hline
     \textbf{G24\_u}&1$>$2,4$>$1,1$>$6,4$>$2,5$>$2,4$>$3,5$>$3,3$>$6,4$>$6,5$>$6\\\hline
     \textbf{G24\_1}&1$>$2,3$>$2,4$>$2,5$>$2,6$>$2,3$>$6,4$>$6,5$>$6\\\hline
     \textbf{G24\_f}&2$>$1,6$>$1,2$>$3,2$>$4,6$>$2,6$>$3,6$>$4,6$>$5\\\hline
     \textbf{G24\_uf}&\\\hline
     \textbf{G24\_2}&1$>$2,4$>$1,5$>$1,1$>$6,3$>$2,4$>$2,5$>$2,4$>$3,3$>$6,4$>$6,5$>$6\\\hline
     \textbf{G24\_2u}&4$>$1,1$>$5,1$>$6,3$>$2,4$>$2,2$>$5,2$>$6,3$>$5,3$>$6,4$>$5,4$>$6\\\hline
     \textbf{G24\_3}&1$>$3,6$>$1,2$>$3,4$>$2,5$>$2,6$>$2,4$>$3,5$>$3,6$>$3,6$>$5\\\hline
     \textbf{G24\_3b}&1$>$2,1$>$3,1$>$6,3$>$2,4$>$2,5$>$2,6$>$2,4$>$3,4$>$5,4$>$6,5$>$6\\\hline
     \textbf{G24\_3f}&2$>$1,5$>$1,6$>$1,2$>$4,6$>$2,6$>$3,5$>$4,6$>$4,6$>$5\\\hline
     \textbf{G24\_4}&1$>$2,1$>$3,1$>$5,1$>$6,3$>$2,4$>$2,6$>$2,3$>$5,4$>$5,4$>$6\\\hline
     \textbf{G24\_5}&1$>$2,4$>$1,5$>$1,1$>$6,3$>$2,4$>$2,5$>$2,4$>$3,3$>$6,4$>$6,5$>$6\\\hline
     \textbf{G24\_6a}&1$>$2,1$>$6,3$>$2,4$>$2,5$>$2,3$>$6,4$>$5,4$>$6,5$>$6\\\hline
     \textbf{G24\_6b}&1$>$2,1$>$6,3$>$2,4$>$2,5$>$2,3$>$6,4$>$5,4$>$6,5$>$6\\\hline
     \textbf{G24\_6c}&1$>$2,1$>$5,1$>$6,3$>$2,4$>$2,5$>$2,4$>$3,3$>$6,4$>$5,4$>$6,5$>$6\\\hline
     \textbf{G24\_6d}&1$>$2,4$>$1,1$>$6,3$>$2,4$>$2,5$>$2,4$>$3,5$>$3,4$>$6,5$>$6\\\hline
     \textbf{G24\_7}&1$>$2,1$>$3,1$>$5,4$>$2,6$>$2,4$>$3,6$>$3,4$>$5,6$>$5\\\hline
     \textbf{G24\_8a}&1$>$3,5$>$1,1$>$6,4$>$2,5$>$2,2$>$6,4$>$3,5$>$3,4$>$6,5$>$6\\\hline
     \textbf{G24\_8b}&1$>$2,4$>$1,5$>$1,3$>$2,4$>$2,5$>$2,5$>$3,3$>$6,4$>$6,5$>$6\\\hline
     \textbf{G24v\_3}&1$>$2,6$>$1,4$>$2,5$>$2,6$>$2,4$>$3,6$>$3,6$>$5\\\hline
     \textbf{G24v\_3b}&1$>$2,1$>$3,4$>$1,4$>$2,5$>$2,6$>$2,4$>$3,5$>$3,6$>$3,4$>$5\\\hline
     \textbf{G24w\_3}&1$>$2,1$>$3,4$>$2,5$>$2,6$>$2,4$>$3,5$>$3,6$>$3\\\hline
     \textbf{G24w\_3b}&1$>$2,1$>$3,4$>$2,5$>$2,6$>$2,4$>$3,5$>$3,6$>$3\\\hline
     \end{tabular}
}
\vspace{-5mm}
\end{table*}

\section{Conclusions and future works}
\vspace{-2mm}
\label{sec:conclusion}
Maintaining and promoting diversity in EAs is crucial to enable them adapt to dynamic environments in DCOPs. 
We have surveyed analysis of the diversity mechanisms, ranging from chaos local search, crowding, fitness diversity, opposition and random immigrants with a base DE algorithm for solving DCOPs. 
We have seen that diversity can be highly beneficial for enhancing the capabilities of DE for solving DCOPs.
We found that diversity mechanisms that are effective for one problem may be ineffective for other problems, and vice versa.
We observed that in some cases, the diversity mechanisms tend to have an opposing force toward the constraint handling technique. To gain more insights, comparing combination of diversity promoting mechanisms with different constraint handling techniques is worth to be more investigated.
Another future study is the application of popular niching methods like clearing, fitness sharing and others with a moving peak benchmark in DCOPs. As these methods show their best effectiveness in multi-modal optimization, comparing them in the current work over our applied benchmark was not fair.

\section*{Acknowledgment}
\vspace{-2mm}
This work has been supported through Australian Research Council (ARC) grant DP160102401.

\bibliographystyle{IEEEtran} 

\bibliography{main.bib}
\end{document}